\documentclass[11pt]{article}

\usepackage[margin=1in]{geometry}
\usepackage{graphicx}
\usepackage{amsmath,amssymb}
\usepackage{xcolor}
\usepackage{setspace}
\usepackage{caption}
\usepackage[colorlinks=true,linkcolor=blue,citecolor=blue,urlcolor=blue]{hyperref}
\usepackage[numbers,sort&compress,super]{natbib}   
\usepackage{titlesec}
\usepackage{algorithm2e}
\usepackage{booktabs}
\usepackage{float}
\usepackage{etoc}
\usepackage{hyperref}
\bibpunct{}{}{,}{s}{}{,}

\IfFileExists{naturemag.bst}{\bibliographystyle{naturemag}}{\bibliographystyle{unsrtnat}}

\DeclareCaptionLabelSeparator{pipe}{ $\vert$ }
\titleformat{\section}{\bfseries\large}{\thesection}{0.6em}{}
\titleformat{\subsection}{\bfseries\normalsize}{\thesubsection}{0.6em}{}
\titlespacing*{\section}{0pt}{1.3em}{0.3em}
\titlespacing*{\subsection}{0pt}{0.9em}{0.2em}

\begin{document}

\begin{center}
{\LARGE\bfseries Stronger artificial intelligence breakthroughs\\[0.15em]
hasten their own exhaustion}\\[1.0em]

{\normalsize Beichen Lu$^{1,2,\dagger}$, Honglin Bao$^{2,3,\dagger,*}$, Qi Huang$^{2}$, Qingguo Meng$^{1}$, James A. Evans$^{2,3,4,5,*}$}\\[0.5em]
{\small\itshape
$^{1}$School of Public Policy and Management, Tsinghua University, Beijing, China.\\
$^{2}$Knowledge Lab, University of Chicago, Chicago, IL, USA.\\
$^{3}$Data Science Institute, University of Chicago, Chicago, IL, USA.\\
$^{4}$Department of Sociology, University of Chicago, Chicago, IL, USA.\\
$^{5}$Santa Fe Institute, Santa Fe, NM, USA.}\\[0.4em]
{\small
$^{\dagger}$These authors contributed equally.\\
$^{*}$Correspondence: \href{mailto:honglinbao@uchicago.edu}{honglinbao@uchicago.edu}; \href{mailto:jevans@uchicago.edu}{jevans@uchicago.edu}}
\end{center}

\vspace{0.6em}

\section*{Abstract}

Progress in artificial intelligence is widely assumed to scale with effort: more researchers, papers and investment should accelerate discovery. Theories of technological growth instead locate the difficulty of discovery in the space of ideas. Here we show that neither scale nor space is enough: breakthroughs depend on diverse attention, which collective concentration can deplete. Across 36 years of computer science, we analyze 7.2 million papers, 73,089 peer reviews and 3.7 million researchers and their topic shifts. The field expanded while its attention narrowed sharply after the Transformer. Over half of topic-shifting researchers moved into AI, especially mid-career ones, and reviewer confidence fell by a full point on a 5-point scale. Each successive breakthrough, from AlexNet to large language models, raised the field's carrying capacity for ideas but was followed by steeper exhaustion of those ideas. Bigger breakthroughs capture attention more completely, depleting the diversity their successors may draw upon. Structural biology shows the same cycle after cryo-EM and AlphaFold. Simulations show how well-intentioned and widely used policies such as reciprocal reviewing empower prolific authors and further narrow what the field accepts. We call for institutions that protect diverse attention in artificial intelligence rather than reward concentration on the winning paradigm.



\vspace{5mm}
Artificial intelligence (AI) has never had more researchers, papers, compute, or investment, and the working assumption of the field is that discovery scales with each. Yet the paradigm now dominating AI spent decades as a minority program that earlier consensus had largely discarded~\cite{lecun2015deep}. Since its return, the field has seen three major breakthroughs in 14 years: (1) AlexNet~\cite{krizhevsky2012imagenet}, the vision model that kicked off the era of deep learning; (2) the Transformer~\cite{vaswani2017attention}, which enabled language classification and translation at scale; and (3) large language models (LLMs), which enabled language generation and complex interaction. Each design was more capable than the last. Each drew the field toward itself more completely than the last. And each was followed by convergence and claims of imminent stagnation~\cite{park2023papers}. If the next breakthrough must be recombined from ideas the current one has not yet absorbed, then a field that converges completely on its winner may be spending the material its successor requires. Whether that is happening is an empirical question current theories of technological progress have not been built to ask.
 
The standard theory of innovation-driven growth is Schumpeter's theory of creative destruction~\cite{schumpeter1942capitalism}, formalized by Aghion and Howitt~\cite{aghion1992model} and recognized in 2025 by the Nobel Prize in Economic Sciences~\cite{nobel2025economics}. The theory models investment in innovation as a path to displace obsolete predecessors, dominate the market, and command high prices as monopoly rents. Technical advances arrive as a stochastic process whose intensity depends on R\&D spending but not the extant field of ideas.
 
That ideas may be getting harder to find is well documented, however. Aggregate research productivity halves roughly every thirteen years~\cite{bloom2020ideas} before a recent boost from large language models~\cite{kusumegi2025scientific, siler2026diffusion}, alongside a doubling of experiments per publication~\cite{shi2015weaving}. Disruptiveness has declined across fields and patent classes~\cite{park2023papers, wu2019large}, teams grow and specialize as the ``burden of knowledge'' accumulates~\cite{jones2009burden}, and attention within growing fields concentrates on increasingly entrenched canons~\cite{chu2021slowed, bhattacharya2020stagnation}. Growth theory has absorbed these observations. Semi-endogenous formulations and those that assume the best ideas have been ``fished out'' first make the arrival rate of innovations decline~\cite{jones1995r, kortum1997research, segerstrom1998endogenous}, reconciling exponential growth in research with steady growth in output.
 
These formalizations share an assumption with consequences for policy: difficulty is a property of the idea space, not the population searching it, and the researcher community only enters the calculus through its size. But cultural evolution shows that this assumption can fail. When rising seas cut Tasmania off from mainland Australia, its people lost bone tools, fishing, and cold-weather clothing over the following millennia. Formal models attribute such losses to a shrinking pool of learners from which skills must be recombined and transmitted~\cite{henrich2004demography, powell2009late, kline2010population, derex2013experimental}, linking loss to small and isolated populations. Modern AI represents the opposite case. It is among the largest and most connected research communities in history, but ten thousand researchers converging on one paradigm do not probe the frontier in ten thousand directions. The effective diversity of search often contracts even as its scale expands~\cite{chu2021slowed, park2023papers,danchev2019centralized}. Recombinant accounts of growth make such ingredients explicit~\cite{weitzman1998recombinant, arthur2009nature, fleming2001recombinant, youn2015invention} but allow the set of seed ideas to grow. Work on lock-in and research strategy describes fields foreclosing alternatives~\cite{arthur1989competing, david1985clio, rzhetsky2015choosing, foster2015tradition} but does not measure what remains for future recombination. Whether a large and growing field can draw down the diversity its next breakthrough requires has not been measured.
 
AI is the natural place to measure it. Its breakthroughs are recent, discrete, and dated, and the field documents itself in public at a scale no other discipline matches. Here we trace the dynamic across 36 years of computer-science research, drawing on 7,180,635 papers, 73,089 peer reviews, 3,707,765 scientists, and 918,862 documented topical migrations, and we confirm that the same cycle appears in structural biology after cryo-EM and AlphaFold. Data, methods, extended results, and full robustness checks are described in the Supplementary Information. Results are consistent across all robustness specifications tested.

\section*{Concentration without precedent}

\begin{figure}[!t]
    \centering
    \includegraphics[width=0.95\textwidth]{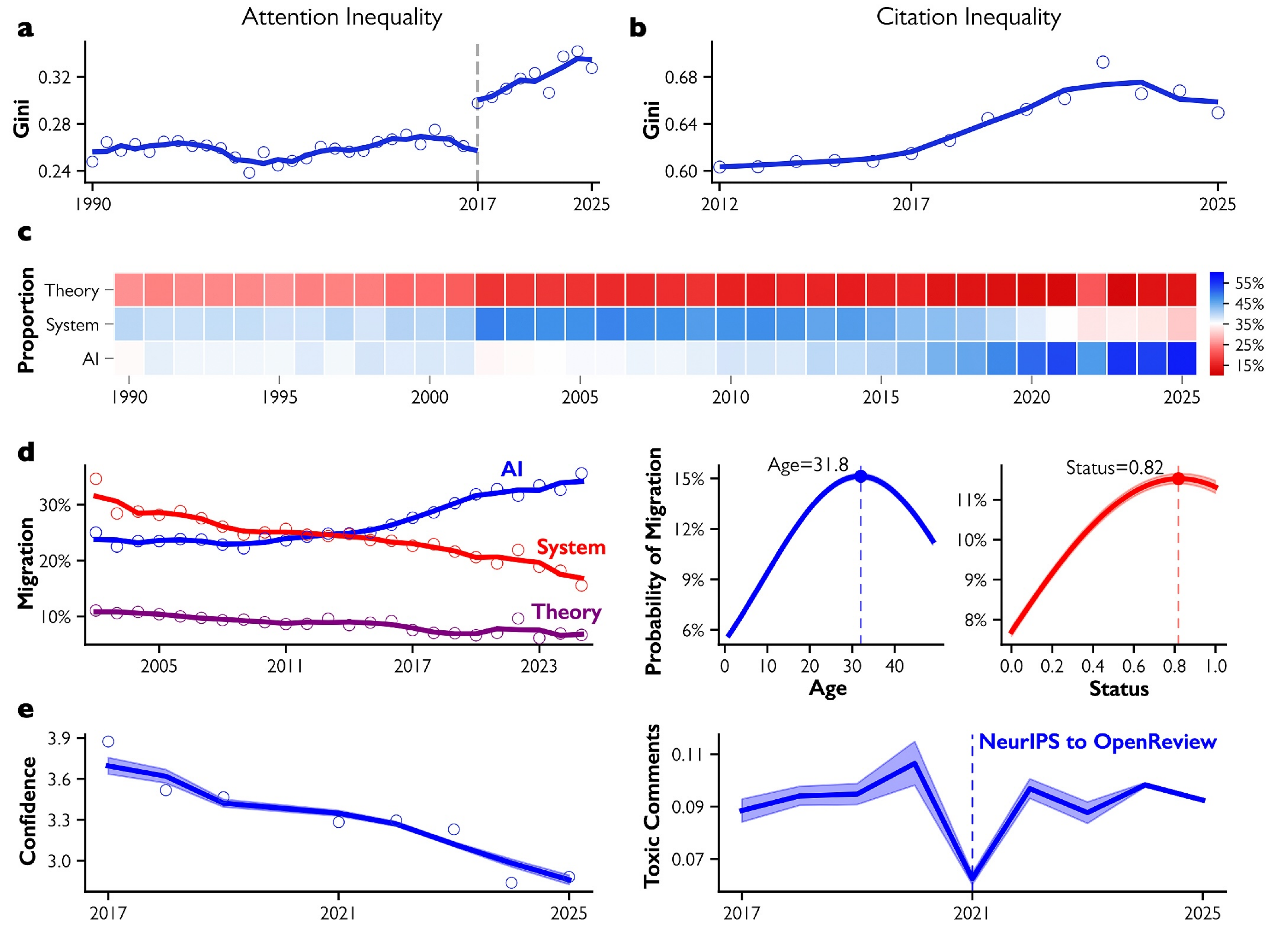}
\caption{\textbf{Attention concentrates.}
    \textbf{a},\textbf{b}, Yearly Gini coefficients of research attention (\textbf{a}) and citations (\textbf{b}) hold flat then break upward in 2017. \textbf{c}, AI's share of computer-science publications increases between 2017 and 2025, while systems and theory contract. \textbf{d}, Author migration probability into AI peaks at age 31.8 and the 82nd yearly publication-rank percentile. \textbf{e}, Across 73{,}089 peer reviews from 2017--2025, reviewer confidence declines from 3.9 to 2.9 on a 0--5 scale (left). Toxic comments increased after 2017, were briefly tempered by the NeurIPS move to OpenReview, and then rebounded (right). Lines in \textbf{a}, \textbf{b}, \textbf{d}, and \textbf{e} are fitted using a three-year rolling average.}
\label{fig:fig1}
\end{figure}

Three converging signatures characterize computer science since 2017, the year the Transformer architecture was introduced (Fig.~\ref{fig:fig1}).

First, attention has narrowed sharply (Fig.~\ref{fig:fig1}a). For each of 7.18 million OpenAlex papers from 1990 to 2025 (Supplementary Section~\ref{sec:emm_corpus}), we extracted the top ten Term Frequency-Inverse Document Frequency (TF-IDF) keywords per abstract and computed the Gini coefficient over annual keyword distributions. We computed inequality analogously over within-year citation flows (Fig.~\ref{fig:fig1}b; Supplementary Section~\ref{sec:app_attention}). Both quantities sit flat and break upward in 2017. The Gini of attention rises from 0.26 to 0.33 within eight years, a 27\% increase relative to a baseline that did not budge through the AI winter, the rise of the Web, the smartphone era, or the deep-learning revolution of 2012. Citation inequality rises from 0.60 to a peak of 0.69 before easing slightly (citation data are limited before 2012). The Transformer disrupted the field on a scale its predecessor did not. The pattern is not an artifact: it survives variation in keyword count $k$ (Supplementary Section~\ref{sec:app_attention_robustness} and Supplementary Fig.~\ref{fig:attention_robustness}). Subfield composition tracks the same break. AI's share of computer-science publications rises from 40\% to over 55\% between 2017 and 2025, while systems research recedes from a long golden age and theory contracts further still (Fig.~\ref{fig:fig1}c).

Second, aggregate concentration is the trace of individual reorientation. We labeled each of 3.7 million authors by the modal subfield of their papers in each active year, then identified migration events as label changes between consecutive active years (Supplementary Section~\ref{sec:app_migration}). Of 918,862 observed transitions, 53.81\% terminate in AI; AI accumulates net inflows in every year of the panel, while systems and theory leak migrants steadily (Fig.~\ref{fig:fig1}d). To identify who migrates, we estimated a logistic regression of ``transition into AI'' on academic age, prior status (within-year publication-count percentile, robust to citation percentile; Supplementary Section~\ref{sec:app_migration_regression}), current productivity, within-field tenure, and unpublished years, with year fixed effects and standard errors clustered at the author level. Both age and prior status enter with strongly significant inverted-U shapes ($p<0.001$ for the squared terms; Fig.~\ref{fig:fig1}d, right; Supplementary Tables~\ref{tab:migration_ai} and~\ref{tab:migration_ai2}). Holding other covariates at their means, the probability of migration peaks at age 31.8 and at the 82nd percentile of prior status. Early-career researchers stay where they trained. Late-career researchers do not retool. Migrants are mid-career, moderately established researchers who leave the work that earned them their standing to pursue the gravitational pull of the field, even at the cost of a pivot penalty~\cite{hill2025pivot}.

Third, evaluation registers the strain. Across 73,089 peer reviews from 26,731 submissions to 45 venues evenly distributed from 2017 to 2025~\cite{bao2025language} (Supplementary Section~\ref{sec:app_peer_review_data}), reviewer confidence declines from 3.9 to 2.9 on a venue-normalized 0--5 scale (Fig.~\ref{fig:fig1}e, left). The decline is sustained, occurs in both AI and non-AI venues, and is consistent with structural difficulty in matching submissions to expertise rather than a venue-specific effect~\cite{beygelzimer2023has}. Three plausible mechanisms compound. Reviewer pools have become flooded with newcomers drawn in by AI's gravity. Reviewers increasingly receive papers laced with AI methods that fall outside their core expertise. Matching submissions to genuinely expert reviewers has become structurally difficult. Acceptance, in this regime, increasingly resembles a coin flip. Toxicity (Supplementary Section~\ref{toxic}) climbs after 2017 (peer-review data on OpenReview are limited before that year), with one striking exception: a sharp drop in 2021, followed by a rebound (Fig.~\ref{fig:fig1}e, right). The timing aligns with 2021, the year when NeurIPS, the largest and most prominent machine learning conference, moved to OpenReview, and the increased visibility appears to have briefly disciplined reviewers' language. The effect, however, did not persist.

\begin{figure}[!t]
    \centering
    \includegraphics[width=1\textwidth]{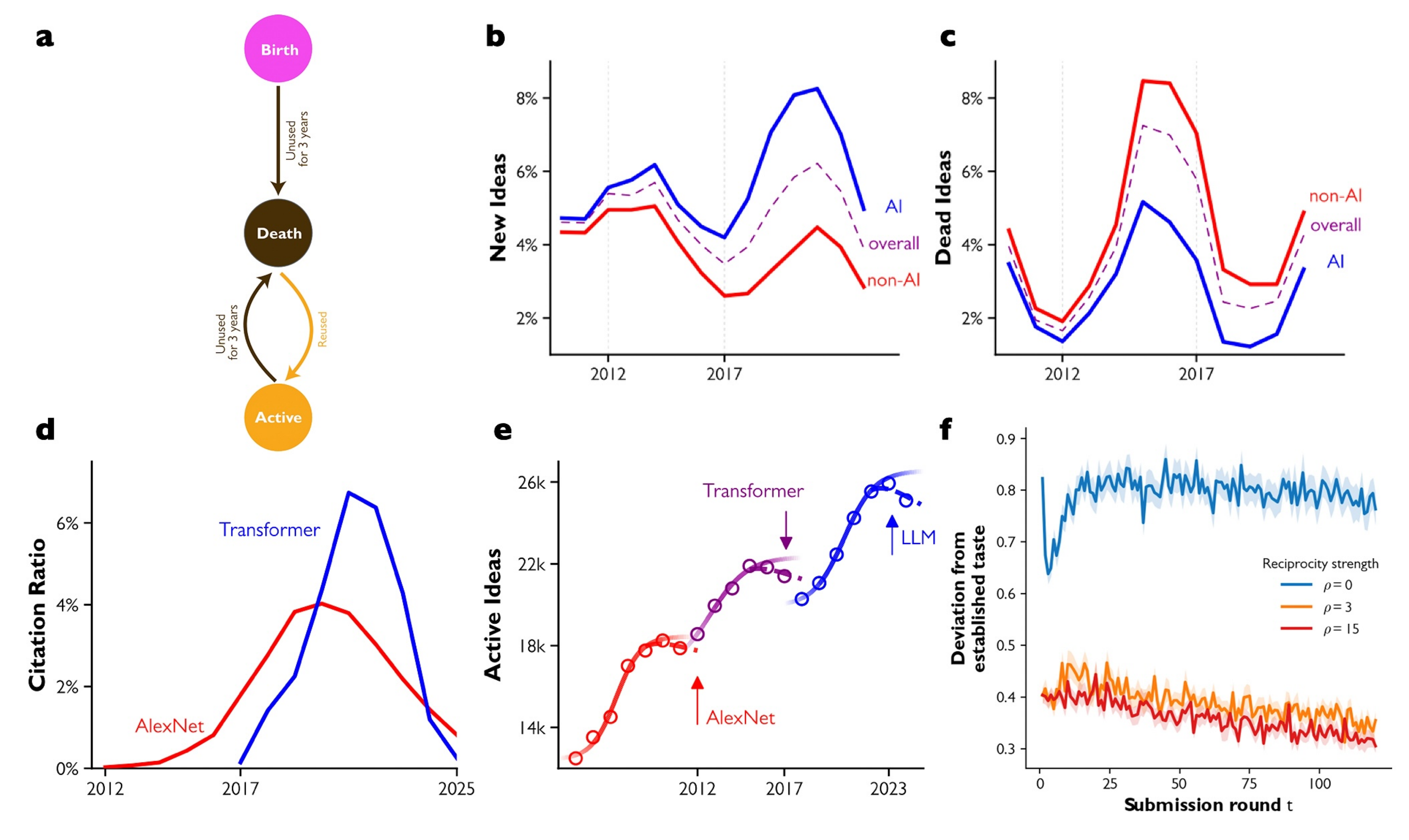}
\caption{\textbf{Cyclical exhaustion and policy backfire.}
    \textbf{a}, Each unique noun in the stable corpus is treated as an idea; ideas are born, die after three years of disuse, and may reactivate. \textbf{b}, The yearly share of newly born ideas rises sharply at AlexNet and the Transformer and then declines. \textbf{c}, Breakthroughs not only birth ideas; they bury them with a lag. \textbf{d}, Citation share for the original AlexNet and Transformer papers peaks and declines for both. Lines in \textbf{b}--\textbf{d} are fitted using a three-year rolling average. \textbf{e}, Active-idea populations fit logistic carrying-capacity curves across three successive cycles (solid lines), with the dashed tails marking \textit{observed} deviation below each modeled ceiling. The drop is modest after AlexNet, severe after the Transformer, and emerges before the new ceiling under LLMs. The cycle is compressing. \textbf{f}, Existing policies may not address this: greater reciprocity strength $\rho$ (high-submission reviewers being asked to review even more) drives stronger convergence toward the tastes of high-submission established researchers.}
    \label{fig:fig2}
\end{figure}

These signatures of the current cycle sit atop an older pattern of \textsc{cyclical exhaustion} (Fig.~\ref{fig:fig2}). To make the cycle visible, we treat each unique and stable noun in abstracts as an idea in the ecosystem, with a birth, death, and possible resurrection (Fig.~\ref{fig:fig2}a; Supplementary Section~\ref{sec:app_idea_ecosystem}). Results are robust across alternative noun-extraction pipelines and noun-phrase representations (Supplementary Sections~\ref{sec:app_idea_robustness} and~\ref{sec:app_text_processing}, Supplementary Fig.~\ref{fig:idea_robustness}). The yearly share of newborn ideas is cyclical, pivoting on the two largest technological disruptions in the dataset (Fig.~\ref{fig:fig2}b: AlexNet 2012; the Transformer 2017). Each breakthrough sharply increases the new-idea rate and is followed by a decay. AI and other fields start out in parallel. With each successive breakthrough, AI's lead in generating new ideas widens and is boosted more by the second wave, but new ideas of other fields drop below an old baseline before 2012. Idea death follows a parallel pattern (Fig.~\ref{fig:fig2}c). Breakthroughs not only birth ideas; they bury them, sometimes with a lag. AI continuously benefits more from these waves than other fields, with more new ideas and fewer dead ones. Breakthroughs are not exempt from their own dynamics, however. They ride the same curve (Fig.~\ref{fig:fig2}d): attention peaks sharply, then drops off, cycle after cycle.

Active idea populations follow logistic ``carrying-capacity'' curves, with each disruption raising the ceiling, where the notion of carrying capacity borrows from ecology~\cite{chapman2018flexible}. At the tail of each cycle, however, populations fall below the ceiling rather than settling on it (Fig.~\ref{fig:fig2}e). The drop is modest after AlexNet, severe after the Transformer, and most striking emerges even before the new ceiling for LLM research, with the active population beginning to bend downward while the post-2022 expansion remains in progress. The post-2022 window is too thin for a definitive claim and the LLM finding awaits additional years of data, but the pattern is consistent with our central argument and inconsistent with the alternative reading that each cycle simply extends its predecessor. Successive cycles in modern AI saturate earlier and exhaust more sharply. The cycle is compressing~\cite{anderson1990technological}. Across 36 years of computer science development, spanning the three major technological advances of AlexNet, the Transformer, and LLMs, we repeatedly observe the same spiral pattern (Supplementary Fig.~\ref{fig:illu}). Notably, this pattern is not confined to computer science. In biology, we observe the same cycle of exhaustion following major non-computational (e.g., cryo-EM) and computational (e.g., AlphaFold) advancements (Supplementary Section~\ref{sec:bio} and Supplementary Fig.~\ref{fig:crossfield_robustness}).

\section*{Policy responses may backfire}

Computer science is not blind to its pathologies. One response now in force across major venues (ICLR, ICML, AISTATS, and ARR) is reciprocal review, in which an author's reviewing load scales with submission count. It is pitched as a remedy for reviewer overload and submission inflation. The intention is sound, but we show that the policy likely narrows the distribution of topics the field can support as an unintended side effect.

Causal identification of this policy from observational data is structurally challenged: the post-policy window is short, and its timing collides with the LLM explosion, which dominates any signal a causal design might detect. We therefore constructed a stylized agent-based model that isolates the mechanism (Supplementary Section~\ref{sim_overall}). Simulated scientists submit and review papers, with reviewer assignment weighted by submission load; the most prolific scientists thereby come to dominate the reviewer pool and, through taste-driven scoring, concentrate the set of accepted papers around their own preferences.

The outcome is consistent across parameter sweeps and 25 random seeds per cell: stronger reciprocity yields faster collapse of identity diversity and tighter clustering of accepted topics around a small set of established tastes (Fig.~\ref{fig:fig2}f). The model is stylized (a two-dimensional topic space and hand-tuned parameters) and we present it not as a prediction but as an illustrative existence proof. The mechanism is nevertheless general: any policy that reweights evaluative authority toward those who produce the most will compress the distribution of ideas the field is able to support.


These findings cash out the conceptual argument with which we began. Creative destruction models the displacement of ideas but not the depletion of the pool they are drawn from; even ``fishing-out'' formulations let that pool only grow, never contract. The assumption fails precisely when destruction is most thorough. A dominant paradigm removes not only its rivals but the diversity from which its own successor would have been recombined. The post-2017 narrowing of attention, the migration of more than half of all field-switching authors into AI, the declining vitality of new ideas, and the steepening tail-decline of active idea populations across successive cycles together describe a system whose creative capacity is being eroded by the success of the dominant paradigms it has produced~\cite{hao2026artificial}. The ``adjacent possible'' has not vanished, but contracted, because the diversity from which adjacent possibilities are drawn has contracted with it~\cite{kauffman2000investigations}.

A Kuhnian reading about paradigm shifts might also suggest that convergence is what normal science looks like, and concentration can be efficient. But our claim concerns what the paradigm leaves behind. A field can press its current paradigm and still retain the variety from which a successor must be assembled, and its institutions decide whether it does. We cannot (yet) predict the next breakthrough in advance~\cite{wu2019large}. Neither bibliometric signals nor peer-review verdicts reliably flag which research trajectories will start the next cycle. That indeterminacy is precisely the case for protecting idea diversity now. The seeds of the next cycle are almost certainly already in the field, but the current incentive structure is built to overlook them~\cite{azoulay2019does}.

The Nobel committee warned that the mechanisms of creative destruction must be upheld if stagnation is to be avoided. Our results identify one such mechanism, intellectual diversity, and show that current institutions are eroding it. The institutions that allocate attention, including reviewing systems, submission fees and rules, reward structures, hiring, promotion, and retirement conventions~\cite{cui2026aging}, all shape whether the next paradigm has room to emerge. Designing them well means resisting the impulse to double down on whichever paradigm is currently winning, taxing concentration rather than exploration, and shifting evaluative authority away from the prolific and toward the novel. Creative destruction sustains growth only as long as it leaves something behind to recombine. A singularity has no successor.


\section*{Methods\label{sec:methods}}
Full details are given in Supplementary Section~\ref{sec:extended_mm}.

\textbf{Bibliometric corpus.} We drew bibliometric data from a December 2025 OpenAlex snapshot containing 463,042,077 work records and applied four filters: (i) primary topic classified as Computer Science; (ii) language English; (iii) publication year 1990--2025; (iv) abstract present. The final corpus contains 7,180,635 papers, 5,299,148 unique authors with at least one computer-science paper, and 3,707,765 authors with at least two papers used in topical-shift analyses. Author disambiguation accuracy was verified at 96\% on a manual 100-author sample (Supplementary Section~\ref{sec:emm_corpus}).

\textbf{Subfield classification.} Papers were assigned to AI, Systems, or Theory using a manually curated concordance from OpenAlex topic identifiers to the standard CSrankings.org taxonomy (full mapping in Supplementary Section~\ref{sec:app_subfield_concordance}). The classification covers 184 OpenAlex topics across the three families.

\textbf{Attention and citation Gini coefficients.} Attention is defined by the important ideas people attend to, rather than all the ideas they study. For each year, we extracted the top $k=10$ TF-IDF keywords per abstract using \textit{scikit-learn}'s \textit{DictVectorizer} and \textit{TfidfTransformer}, aggregated into a year-level keyword-frequency distribution, and computed the Gini coefficient over the resulting distribution. Citation Gini coefficients were computed via a histogram-based Lorenz construction over yearly citation flows (Supplementary Section~\ref{sec:app_attention}). Robustness checks confirm the 2017 break for $k \in \{5,10,15\}$ (Supplementary Section~\ref{sec:app_attention_robustness} and Supplementary Fig.~\ref{fig:attention_robustness}).

\textbf{Idea life cycle.} We use NLTK and spaCy noun-extraction pipelines to extract all ideas (results robust to either, and to noun-phrase representations; Supplementary Sections~\ref{sec:app_idea_robustness} and~\ref{sec:app_text_processing}, Supplementary Fig.~\ref{fig:idea_robustness}) by conceptualizing each unique noun as an idea. An idea is \textit{stable} in year $y$ if it appears in at least 10 papers in years $y$ and $y{+}1$ (to remove arbitrary use). New ideas are stable ideas that have never previously appeared. An idea is \textit{dead} after three consecutive years of absence from stable status; a dead idea is reactivated by reappearance. The active-idea stock is the cumulative set of stable ideas net of currently dead ones (Supplementary Section~\ref{sec:app_idea_ecosystem}).

\textbf{S-shape carrying-capacity model.} For each cycle (pre-AlexNet, AlexNet, the Transformer), we fit the active-idea trajectory to a logistic carrying-capacity model with an exponential tail decay, estimating parameters separately per cycle by nonlinear least squares (Supplementary Section~\ref{sec:app_idea_ecosystem}).

\textbf{Author migration regression.} The author-by-year panel was constructed from authors with at least two papers. The dependent variable is migration into AI, modeled as a logistic regression of academic age, prior status, their squared terms, current-year publication percentile, within-field tenure, unpublished years, and year fixed effects. Standard errors are clustered at the author level (Supplementary Section~\ref{sec:app_migration}). Inverted-U coefficients on age and prior status are stable across alternative status measures (publications versus citations; Supplementary Section~\ref{sec:app_migration_regression}, Supplementary Tables~\ref{tab:migration_ai} and~\ref{tab:migration_ai2}).

\textbf{Peer-review confidence and toxicity.} Reviews were drawn from 45 venues 2017--2025 via OpenReview~\cite{bao2025language} (Supplementary Section~\ref{sec:app_peer_review_data}). Confidence scores were min--max normalized to a 0--5 scale per venue. Toxicity was scored using the ParaDetox classifier~\cite{logacheva2022paradetox} (style accuracy 0.89, AUC--ROC $>$ 0.98; Supplementary Section~\ref{toxic}).

\textbf{Agent-based simulations.} The reciprocal-review model embeds $N=100$ scientists ($N_P=5$ prolific, $\mathrm{Poisson}(20)$ submission rate; 95 regular, $\mathrm{Poisson}(2)$) over $T=120$ rounds. Reviewer assignment weights $w = 1 + \rho k$ couple reviewing load $w$ to submission load $k$, with demonstrative reciprocity strength $\rho \in \{0, 3, 15\}$. An author's reviewing taste is shaped by their own research identity, which is updated by the papers they produce, with accepted papers weighted more heavily. Each simulation cell uses 25 seeds; full algorithms and parameter tables are in Supplementary Section~\ref{sim_overall}.



\section*{Data availability}
The bibliometric corpus derives from the December 2025 OpenAlex snapshot and the peer-review corpus from OpenReview, both publicly accessible. Processed data supporting the findings of this study are deposited in a public \href{https://github.com/Honglin-Bao-Chicago/vitality_collapse}{repository}.

\section*{Code availability}
All code for data processing, analysis, and the agent-based simulations is deposited in a public \href{https://github.com/Honglin-Bao-Chicago/vitality_collapse}{repository}.

\section*{Acknowledgements}
We thank Knowledge Lab members for valuable feedback. Computation used the University of Chicago Midway / Data Science Institute Research Computing Cluster.

\section*{Author contributions}
B.L. and H.B. co-led the project and contributed equally. H.B., B.L., and J.A.E. designed the study. B.L. prepared the data. H.B. and B.L. conducted the analysis and drafted the paper. H.B. and Q.H. designed and implemented the simulations. J.A.E. revised the manuscript. Q.M. and J.A.E. supervised the work.

\section*{Competing interests}
J.A.E. is affiliated with Google. The remaining authors declare no competing interests.

\section*{Additional information}
\textbf{Supplementary information} is available for this paper: Supplementary Section~\ref{sec:extended_mm} (extended data and methods); Supplementary Section~\ref{sec:robustness} (robustness across measures of attention concentration, idea-extraction methods, and author-migration results under alternative status measures); Supplementary Section~\ref{sec:si_figures_tables} (Supplementary Figures and Tables).

\textbf{Correspondence and requests for materials} should be addressed to Honglin Bao or James A. Evans.

\bibliography{references_policyforum}


\clearpage

\clearpage

\clearpage

%

\appendix

\begin{center}
{\Huge \textbf{Supplementary Information}}\\[1em]
{\large Stronger artificial intelligence breakthroughs hasten their own exhaustion}\\[0.6em]
{\normalsize Beichen Lu, Honglin Bao, Qi Huang, Qingguo Meng, James A. Evans}
\end{center}
\vspace{1em}

\tableofcontents

\newpage


\setcounter{secnumdepth}{3}
\renewcommand{\thesection}{\arabic{section}}
\setcounter{section}{0}

\setcounter{figure}{0}
\renewcommand{\thefigure}{\arabic{figure}}
\setcounter{table}{0}
\renewcommand{\thetable}{\arabic{table}}
\captionsetup[figure]{labelformat=simple,name={Supplementary Fig.},labelsep=pipe,labelfont=bf,
                      font=small,justification=raggedright,singlelinecheck=false}
\captionsetup[table]{labelformat=simple,name={Supplementary Table},labelsep=pipe,labelfont=bf}

\section{Methodology Details}
\label{sec:extended_mm}

\subsection{Bibliometric Corpus}
\label{sec:emm_corpus}

We draw our bibliometric data from OpenAlex, a large-scale open catalog of scholarly works, authors, institutions, and citation links. We downloaded a snapshot of the database comprising 463,042,077 work records in total. We applied the following filters to each paper to select the valid ones for our research: (i) primary topic is classified as Computer Science; (ii) language is English; (iii) publication year falls within the range 1990--2025; and (iv) the abstract exists. After applying these criteria, the final dataset contains 7,180,635 computer science papers. Most of the analysis is completed based on this processed dataset, while additional filtering may be required to meet the requirements of specific algorithms.

We scan the full OpenAlex snapshot to build a dataset of unique author identifiers. This yields 75,772,839 distinct authors. Among these authors, 5,299,148 appear in our filtered computer science dataset. For their topical shift behaviors we keep authors with at least 2 papers (3,707,765 authors).

Name disambiguation is always a concern in large-scale data for scientific papers especially in computer science where there are a lot of Chinese names. To validate the author matching, we randomly sampled 100 author names and manually checked for erroneous merges between distinct authors sharing identical names (i.e., whether their papers indeed belong to them). This robustness check yielded a 96\% accuracy rate. A match means that the papers attributed to the author in the OpenAlex data are, with certainty, their own.

We use the \texttt{counts\_by\_year} field provided by OpenAlex, which records the number of citations each paper received in each calendar year\footnote{Note that this entry provides incomplete citation records for earlier works, particularly for citations that occurred before 2012. Thus in Fig. 1 of the main text we did not show citation dynamics before 2012.}.


\subsection{Idea Extraction}
\label{sec:app_text_processing}

We reconstruct the full abstract text for each paper from the metadata provided by OpenAlex and concatenate it with the paper title to form a single text passage per work. We extract ideas as nouns from the title and abstract of each paper, using two independent natural language processing (NLP) pipelines. The first uses NLTK with the part-of-speech tagger and WordNet lemmatizer; the second uses spaCy with language model \texttt{en\_core\_web\_lg}. Both pipelines tokenize the passage, identify nouns, lemmatize them, and filter out uninformative tokens (including stopwords, punctuation, numerals, and single-character strings). Running both pipelines in parallel allows us to verify that downstream results are not sensitive to the choice of NLP toolchain. We further tested representing ideas using phrases (e.g., multiple nouns), and all conclusions remain robust (see Supplementary Section~\ref{sec:app_idea_robustness}).


\subsection{Peer Review Data}
\label{sec:app_peer_review_data}

The open review data set was mainly adopted from \citep{bao2025language}. It includes 26,731 submissions to 45 conferences from 2017 to 2025 (before 2017 the OpenReview data is very limited): 1st ContinualAI Unconference, AAAI Conference on Artificial Intelligence 2024 and 2025, Symposium on Advances in Approximate Bayesian Inference 2024, Conference on Language Modeling 2024, Cooking Robotics Workshop 2024, Conference on Robot Learning 2023 and 2024, Conference on Parsimony and Learning 2024, IEEE/CVF Conference on Computer Vision and Pattern Recognition 2023 and 2024, Workshop on Distributed Infrastructure for Common Good 2023, Workshop on Embodiment-Aware Robot Learning 2024, European Conference on Computer Vision 2024, Conference on Empirical Methods in Natural Language Processing 2023, European Space Power Conference 2023, Fast, Low-resource, and Accurate Organ and Pan-cancer Segmentation in Abdomen CT 2023, SIGIR Workshop on Generative Information Retrieval 2024, ACM/IEEE International Conference on Human-Robot Interaction 2023 and 2024, International Conference on Integration of Science and Technology for Sustainable Development 2024, International Conference on Learning Representations 2017 to 2025, International Conference on Machine Learning 2023 and 2024, ACM International Conference on the Theory of Information Retrieval 2024, International Joint Conference on Artificial Intelligence 2024, International Semantic Web Conference 2024, ACM SIGKDD Conference on Knowledge Discovery and Data Mining 2023 and 2024, ACM International Conference on Multimedia 2024, Conference on Neural Information Processing Systems 2021-2024, Neuro-Symbolic Learning and Reasoning in the Era of Large Language Models 2024, Next-generation Data Governance Workshop 2024, Tsinghua University Advanced Machine Learning 2024.


\subsection{Attention and Citation Inequality}
\label{sec:app_attention}

To measure what important topics researchers attend to (rather than all ideas that appear), we extract the \textit{most prominent} keywords from each paper using term frequency--inverse document frequency (TF-IDF) weighting and aggregate the results into a year-level keyword frequency distribution. The procedure is applied independently within each calendar year.

The TF-IDF computation is implemented via \texttt{DictVectorizer} and \texttt{TfidfTransformer} from the Python package scikit-learn. The output $C_y$ is a frequency distribution over the vocabulary $\mathcal{V}^{\mathrm{kw}}_y$: each entry $C_y[w]$ records the number of papers in year $y$ for which term $w$ ranks among the top-$k$ keywords. The conclusions are robust to different $k$ (Supplementary Section~\ref{sec:app_attention_robustness}).

We quantify the concentration of research attention by computing the Gini coefficient over this distribution. Let $x_1 \leq x_2 \leq \cdots \leq x_n$ denote the sorted nonzero values of $C_y$. The Gini coefficient for year $y$ is:

\begin{equation}
G_y = \frac{2 \sum_{i=1}^{n} i \cdot x_i}{n \sum_{i=1}^{n} x_i} - \frac{n+1}{n}.
\end{equation}

A value of zero indicates perfect equality, while values approaching one indicate extreme concentration.

We measure citation inequality by computing the Gini coefficient of the citation distribution in each calendar year. Considering the scale of the data, we use a histogram-based Lorenz curve method to calculate the Gini coefficient.

To examine how citations flow to computer science papers in each year, for each year $y$, let $\mathcal{P}_y$ represent the set of all cited papers in year $y$. For each paper $p \in \mathcal{P}_y$, let $c_p^{(y)}$ denote the number of citations paper $p$ received in year $y$. We construct a citation histogram $H_y$:

\begin{equation}
H_y[j] = \left|\left\{p \in \mathcal{P}_y : c_p^{(y)} = j\right\}\right|, \quad j = 1, 2, 3, \ldots
\end{equation}

Let $\{(j_1, H_y[j_1]),\, \ldots,\, (j_m, H_y[j_m])\}$ denote the histogram entries, sorted in ascending order $j_1 < j_2 < \cdots < j_m$. Define the total citations:

\begin{equation}
Z_y = \sum_{\ell=1}^{m} j_\ell \cdot H_y[j_\ell].
\end{equation}

The cumulative population share and cumulative citation share after the $\ell$-th bin are:

\begin{equation}
Q_\ell = \frac{1}{|\mathcal{P}_y|}\sum_{u=1}^{\ell} H_y[j_u], \qquad
L_\ell = \frac{1}{Z_y}\sum_{u=1}^{\ell} j_u \cdot H_y[j_u],
\end{equation}

with $Q_0 = L_0 = 0$. The area under the Lorenz curve is estimated by the trapezoidal rule:

\begin{equation}
A = \sum_{\ell=1}^{m} \frac{L_{\ell-1} + L_\ell}{2} \cdot (Q_\ell - Q_{\ell-1}).
\end{equation}

The Gini coefficient for year $y$ is:

\begin{equation}
G_y = 1 - 2A.
\end{equation}


\subsection{Toxicity}
\label{toxic}

The ParaDetox framework \cite{logacheva2022paradetox}, a BERT-family model, provides a reliable basis for labeling toxic peer review comments by defining toxicity as a controlled style dimension and enforcing it through a rigorous multi-stage annotation pipeline. Each example uses independent crowdsourced tasks, with labels aggregated via Dawid--Skene and filtered at high confidence ($\geq 90\%$), which substantially reduces noise and ambiguity. Empirically, this definition of toxicity yields high accuracy: models trained on ParaDetox achieve up to $0.89$ style accuracy in automatic evaluation, an AUC--ROC above $0.98$ outperforming unsupervised baselines, and $83\%$ accuracy under human evaluation. When applying it, texts whose token length exceeds 510 (i.e., the maximum context window - 2) are partitioned into overlapping windows of 510 tokens using a stride of 256, yielding an overlap of 254 tokens between consecutive windows. After prepending and appending the requisite special tokens, each window is scored independently by the classifier. Full-review-comment-level toxicity is then obtained by taking the maximum toxicity probability across all windows, such that a comment is labeled toxic if any of its constituent windows is classified as toxic. Texts of 510 tokens or fewer yield a single window and are processed without sliding.


\subsection{Author Migration}
\label{sec:app_migration}

\paragraph{\textbf{Identification of Migration Events.}}
Based on the subfield classification described in Supplementary Section~\ref{sec:app_subfield_concordance}, we count the number of papers each author publishes in each of the three identifiable subfields (AI, System, and Theory) per year. If author $i$'s publications in year $y$ are dominated by one of the three identifiable subfields, that author's field label for year $y$ is assigned accordingly:

\begin{equation}
\ell_{i,y} = \arg\max_{f \,\in\, \{\text{AI},\, \text{System},\, \text{Theory}\}} \; a_{i,y}^{f},
\end{equation}

where $a_{i,y}^{f}$ denotes the number of papers author $i$ published in subfield $f$ during year $y$.

This procedure yields a \emph{subfield timeline} for each author: the ordered sequence of (year, label) pairs for all years in which the author has a valid label. Let $y_1 < y_2 < \cdots < y_{T_i}$ denote author $i$'s active years. A \textbf{migration event} occurs at time step $t$ ($t \geq 2$) whenever the label changes between consecutive active years. We represent each migration event as a tuple:

\begin{equation}
e_{i,t} = \left(\ell_{i,\,y_{t-1}},\;\; \ell_{i,\,y_t}\right), \qquad \text{where } \ell_{i,\,y_{t-1}} \neq \ell_{i,\,y_t}.
\end{equation}

In addition to migration events, we derive the following variables for each author-year observation:

\begin{enumerate}
    \item \textbf{Academic age} $\alpha_{i,y}$: defined as $\alpha_{i,y} = y - y_i^{\text{first}}$, where $y_i^{\text{first}}$ is author $i$'s first publication year across the full OpenAlex corpus.
    \item \textbf{Current-year publication count} $n_{i,y}$: the total number of papers author $i$ published in year $y$.
    \item \textbf{Cumulative publications} $N_{i,y}$: the total number of papers author $i$ has published in all years strictly before $y$, $N_{i,y} = \sum_{y' < y} n_{i,y'}$.
    \item \textbf{Cumulative citations} $C_{i,y}$: the total number of citations author $i$ has received in all years strictly before $y$, $C_{i,y} = \sum_{y' < y} c_{i,y'}$.
    \item \textbf{Prior status $s_{i,y}$}: We use the yearly average publication/citation percentile rank before year $y$ as a measure of prior status of author $i$ in year $y$, as the distribution is highly long-tailed and this approach avoids confounding cumulative citations and publications with academic age.
    \item \textbf{Tenure in the current field} $h_{i,y}$: for $y = y_t$, the number of consecutive active years the author has spent in the current subfield prior to year $y_t$. Tenure resets to zero upon a migration event.
    \item \textbf{Gap years} $g_{i,y}$: for $y = y_t$, the number of inactive years (years without any papers) between consecutive active years, $g_{i,y} = y_t - y_{t-1} - 1$.
\end{enumerate}

\paragraph{\textbf{Inflow Rate.}}
For each year $y$, let $\mathcal{A}_y$ denote the set of authors with a valid label in year $y$. Writing $y_t$ for author $i$'s $t$-th active year, let $t(i,y)$ be the index with $y_{t(i,y)} = y$. We define the \textbf{inflow count} into subfield $f$ in year $y$ as:

\begin{equation}
M_y^{f} = \left|\left\{i \in \mathcal{A}_y \;:\; \ell_{i,y} = f \;\;\text{ and }\;\; \Big( t(i,y) = 1 \;\;\text{ or }\;\; \ell_{i,\,y_{t(i,y)-1}} \neq f \Big)\right\}\right|,
\end{equation}

where $t(i,y) = 1$ indicates that year $y$ is the author's first active year (i.e., a new entrant). That is, $M_y^{f}$ counts both researchers who migrate from another subfield into $f$ and new entrants whose first valid label is $f$. The \textbf{inflow rate} into subfield $f$ is defined by the rate of migrated authors over all authors in the field (including new authors):

\begin{equation}
r_y^{f} = \frac{M_y^{f}}{|\mathcal{A}_y|}.
\end{equation}

We compute $r_y^{f}$ for $f \in \{\text{AI},\, \text{System},\, \text{Theory}\}$ and report the three time series in the paper, as shown in Fig. 1 panel d.

\paragraph{\textbf{Age, Status, and Migration Behavior.}}
We further investigate the relationship between academic age, status, and migration behavior through a panel regression analysis.

\textit{Panel Construction.} For the regression analysis, we collapse the three-way subfield labels into a binary classification: AI ($\ell = 1$) versus Non-AI ($\ell = 0$, combining System and Theory). If a given author-year observation has more AI papers than non-AI papers, the label is 1; if the reverse holds, the label is 0. We construct an author$\times$year panel by iterating over each author's subfield timeline and retaining observations for which a lagged label is available. The dependent variable is $\texttt{migrate\_to\_ai}_{i,y} = 1$ if author $i$'s label transitions from 0 to 1 (Non-AI to AI) between the previous active year and year $y$, and 0 otherwise.

\textit{Model Specification.} The baseline specification is a logit model:
\begin{equation}
\Pr\!\left(\texttt{migrate\_to\_ai}_{i,y} = 1\right) = \Lambda\!\left(\beta_0 + \beta_1 \, \alpha_{i,y} + \beta_2 \, \alpha_{i,y}^2 + \beta_3 \, s_{i,y} + \beta_4 \, s_{i,y}^2 + \mathbf{X}_{i,y}'\boldsymbol{\gamma} + \delta_y\right),
\end{equation}
where $\Lambda(\cdot)$ is the logistic function, $\alpha_{i,y}$ is academic age, $s_{i,y}$ is prior status, and $\delta_y$ denotes year fixed effects. Academic age and prior status each enter as a quadratic to capture potential nonlinearity (we also have models without quadratic terms, as shown in Supplementary Section~\ref{sec:app_migration_regression}). The control vector $\mathbf{X}_{i,y}$ includes:
\begin{equation}
\mathbf{X}_{i,y} = \Big(\mathrm{pct}(n_{i,y}),\;h_{i,y},\;\; g_{i,y}\Big)',
\end{equation}
where $\mathrm{pct}(\cdot)$ denotes the percentile rank of the argument within the focal year, $n_{i,y}$ is the current-year publication count, $h_{i,y}$ is within-field tenure, and $g_{i,y}$ is gap years. Standard errors are clustered at the author level.


\subsection{Scientific Ecosystem and Idea Lifecycle}
\label{sec:app_idea_ecosystem}

\paragraph{\textbf{Idea Lifecycle Definitions.}}
Using the idea extraction pipeline described above (which is also robust to different idea extraction NLP pipelines and another representation of ideas using phrases, Supplementary Section~\ref{sec:app_idea_robustness}), we first conceptualize each unique noun as an individual idea within the scientific ecosystem. The birth, persistence, death, and reactivation of these ideas across years collectively constitute an ecology that shows the dynamics of scientific progress.

To suppress noise, we define a set of stable ideas for each year. Let $\mathrm{df}(w, y)$ denote the number of papers in which idea $w$ appears in year $y$. An idea $w$ is considered stable in year $y$ if it satisfies:

\begin{equation}
\mathrm{df}(w, y) \geq 10 \quad \text{and} \quad \mathrm{df}(w, y+1) \geq 10.
\end{equation}

This criterion requires an idea not only to be present in the current year but also to maintain a stable presence in the following year, thereby filtering out ephemeral terminology and arbitrary uses (i.e., fewer than ten occurrences across millions of papers).

Let $\mathcal{S}_y$ denote the resulting set of stable ideas in year $y$. Building on the annual stable idea set, we define three lifecycle categories that together characterize the ecological dynamics.

\textbf{New ideas.} An idea is new in year $y$ if it enters the stable set for the first time:

\begin{equation}
\mathcal{B}_y = \left\{w \in \mathcal{S}_y : w \notin \bigcup_{y'=1990}^{y-1} \mathcal{S}_{y'}\right\}.
\end{equation}

\textbf{Dead ideas.} An idea is considered dead in year $y$ if it was previously a member of the stable set but has been absent for three consecutive years:

\begin{equation}
w \in \mathcal{D}_y \iff w \in \bigcup_{y' < y-2} \mathcal{S}_{y'} \;\;\text{ and }\;\; w \notin \mathcal{S}_{y''},\;\; \forall\, y'' \in \{y-2,\, y-1,\, y\}.
\end{equation}

A dead idea can be reactivated if it reappears as a stable idea in a subsequent year.

\textbf{Active ideas.} The stock of active ideas in year $y$ is the cumulative count of all ideas that have ever been stable, minus those currently dead:

\begin{equation}
\mathcal{V}_y = \left(\bigcup_{y'=1990}^{y} \mathcal{S}_{y'}\right) \;\setminus\; \mathcal{D}_y.
\end{equation}

From these definitions we derive two time series reported in the paper: the ratio of new ideas $|\mathcal{B}_y| / |\mathcal{S}_y|$, which captures the rate at which novel concepts enter the field, and the active idea count $|\mathcal{V}_y|$, which captures the cumulative dynamics of the scientific ecosystem.

\paragraph{\textbf{Ecological Growth Model.}}
To capture the cyclical dynamics of scientific innovation, we develop a multi-period growth model. Rooted in ecological theory, this model traces the evolution of scientific innovation as a process governed by environmental carrying capacity constraints. We partition the timeline according to the emergence of two breakthrough technologies, AlexNet and the Transformer architecture, into three cycles: before 2012, 2012--2017, and after 2017. Within each cycle, we fit the growth trajectory of the active-idea count $|\mathcal{V}_y|$ as defined above using a logistic function with an exponential decline term.

The logistic model describes an S-shaped growth curve that characterizes population expansion under carrying capacity constraints in ecological systems. The mathematical form is

\begin{equation}
L(t) = n_0 + \frac{K}{1 + e^{-r(t - t_0)}},
\end{equation}

where $n_0$ is the lower asymptote of the curve (the baseline level inherited from the preceding cycle, approached as $t$ falls well below $t_0$), $n_0 + K$ is the \emph{carrying capacity} that determines the upper bound of idea accumulation, $r$ is the intrinsic growth rate that governs how rapidly new ideas enter the ecosystem, and $t_0$ is the inflection point at which growth transitions from accelerating to decelerating.

In each cycle, the observed trajectory eventually deviates below the idealized logistic curve as crowding-out pressures set in. Writing $\hat{V}(t)$ for the fitted value of the active-idea count $|\mathcal{V}_y|$ at $t = y$, we model this deviation with an exponential decay factor applied beyond a saturation year $t_d$:

\begin{equation}
\hat{V}(t) =
\begin{cases}
L(t), & t \leq t_d, \\[4pt]
L(t) \cdot e^{-r_d(t - t_d)}, & t > t_d,
\end{cases}
\end{equation}

where $r_d > 0$ is the decline rate. The divergence between $L(t)$ and $\hat{V}(t)$ for $t > t_d$ quantifies the vitality loss within each paradigm cycle. For each of the three cycles, we estimate the parameters $K$, $r$, $t_0$, $n_0$, $t_d$, and $r_d$ independently by nonlinear least squares via \texttt{scipy.optimize.curve\_fit}.

\paragraph{\textbf{The Stylized Model of Idea Evolution}}
We illustrate a stylized growth model of idea evolution in Supplementary Fig.~\ref{fig:illu}, comprising four stages: \emph{initiation}, where a new paradigm has just emerged; \emph{growth}, where the S-curve reaches its steepest slope and new ideas proliferate; \emph{saturation}, where the idea population calcifies and may even deteriorate; and finally \emph{breakthrough}, a largely unpredictable transformation that lifts the carrying capacity and restarts the cycle.

\subsection{Agent-Based Simulations}
\label{sim_overall}


We model peer review as a paper--reviewer matching and scoring process, where researchers' own taste impacts how they evaluate.

\paragraph{\textbf{Topics and Identities.}}
Scientific topics are embedded in a continuous $d$-dimensional space $\mathcal{X}=\mathbb{R}^d$. Distance in topic space represents conceptual dissimilarity.

Each scientist $i=1,\dots,N$ is characterized at time $t$ by a research identity $\theta_{i,t}\in\mathbb{R}^d$ and a review taste $\tau_{i,t}\in\mathbb{R}^d$. Initial identities are drawn from a prior, $\theta_{i,0}\sim\mathcal{N}(0,\sigma_0^2 I_d)$, and initial tastes are set equal to initial identities, $\tau_{i,0}=\theta_{i,0}$; the two-type specification of $\sigma_0$ used in our simulations is given in the Implementation paragraph below.

\paragraph{\textbf{Submissions.}}
In submission round $t$, scientist $i$ submits $k_{i,t}\in\{0,1,2,\dots\}$ papers, drawn as $k_{i,t}\sim\mathrm{Poisson}(\mu_i)$, where the rate $\mu_i$ is heterogeneous across scientists and fixed over time for simplicity. Heterogeneity in $\mu_i$ is the exogenous source of asymmetry that the evaluation mechanism will amplify.

Each paper $p$ has a topic $x_p\in\mathbb{R}^d$ drawn around the author's current research identity,
\begin{equation}
x_p \sim \mathcal{N}(\theta_{i,t}, \sigma_x^2 I_d),
\end{equation}
where $\sigma_x^2$ captures within-author topical variability. This captures the idea that researchers produce work aligned with their existing expertise, while allowing stochastic exploration.

\paragraph{\textbf{Reviewer Assignment.}}
Each submitted paper is assigned to $R\ge 1$ reviewers. Let $\mathcal{R}(p)$ denote the set of reviewers assigned to paper $p$, and let $i(p)$ denote the author of $p$. Reviewers are sampled from the set $\{1,\dots,N\}\setminus\{i(p)\}$, with selection weights
\begin{equation}
\psi_{j,p,t}
\;\propto\;
\exp\!\left(-\delta \|x_p - \tau_{j,t}\|^2\right)
\cdot
w_{j,t},
\qquad j \neq i(p),
\label{eq:reviewer_weight}
\end{equation}
where $\delta>0$ governs the strength of topical matching and $w_{j,t}>0$ captures reviewer availability. Concretely, the $R$ reviewers are drawn sequentially, each time in proportion to \eqref{eq:reviewer_weight} restricted to scientists not yet assigned to $p$.

Reviewer availability increases with a scientist's own submission volume, which is the mechanical sense in which review is \emph{reciprocal} in this model:
\begin{equation}
w_{j,t} = 1 + \rho\, k_{j,t},
\qquad \rho\ge 0.
\end{equation}
The parameter $\rho$ controls the strength of reciprocity: $\rho=0$ recovers uniform availability, while larger $\rho$ ties reviewing load more tightly to submission load. Per-author submission caps already address the case where a small number of authors submit many papers while the review load cannot scale proportionally (e.g., AAAI, IJCAI, and CVPR).

\paragraph{\textbf{Review Scores.}}
Each reviewer $j\in\mathcal{R}(p)$ provides a scalar score
\begin{equation}
s_{jp}
=
\bar{s}
-
\gamma \|x_p - \theta_{i(p),t}\|^2
-
\beta \|x_p - \tau_{j,t}\|^2
+
\varepsilon_{jp},
\qquad
\varepsilon_{jp} \sim \mathcal{N}(0,\sigma^2),
\end{equation}
where $i(p)$ denotes the author of paper $p$, and $\gamma,\beta>0$. The score decomposes into three components with distinct interpretations:

\begin{itemize}
\item \textbf{Quality} ($-\gamma \|x_p - \theta_{i(p),t}\|^2$). Papers closer to the author's own research identity are of higher quality, reflecting the idea that scientists produce the most solid work within their area of expertise. 

\item \textbf{Taste match} ($-\beta \|x_p - \tau_{j,t}\|^2$). Reviewers additionally penalize papers distant from their own evaluative taste. This term is \emph{reviewer-specific} and is the channel through which reviewer heterogeneity enters the scoring process.

\item \textbf{Idiosyncratic noise} ($\varepsilon_{jp}$). Residual variation is independent of both quality and taste.
\end{itemize}

The constant $\bar{s}$ is the expected score a reviewer would assign to a paper that is both perfectly within its author's expertise ($x_p=\theta_{i(p),t}$) and perfectly aligned with the reviewer's taste ($x_p=\tau_{j,t}$).

The ratio $\gamma/\beta$ governs how much peer review rewards intrinsic quality relative to taste conformity. When $\gamma \gg \beta$, reviewers act primarily as quality evaluators and the system is close to merit-based; when $\gamma \ll \beta$, taste dominates and reviewer identity drives acceptance.

\paragraph{\textbf{Acceptance Decision.}}
The final score of paper $p$ is the average reviewer score,
\begin{equation}
S_p
=
\frac{1}{R}
\sum_{j\in\mathcal{R}(p)} s_{jp}.
\end{equation}
Acceptance is based on relative ranking. Let $\pi = 0.25$ denote the target acceptance rate and let $\bar{S}_t$ be the $(1-\pi)$-quantile of $\{S_p : p \in \mathcal{P}_t\}$. Paper $p$ is accepted if
\begin{equation}
\mathrm{Accept}_p
=
\mathbf{1}\{S_p \ge \bar{S}_t\}.
\end{equation}
Ties at the quantile are broken uniformly at random, so the realized acceptance rate equals $\pi$ up to integer rounding.

\paragraph{\textbf{Identity Evolution.}}
Research identities evolve through production, with selective reinforcement by peer review. Let $\mathcal{P}_{i,t}$ denote the papers submitted by scientist $i$ in round $t$ and let $\mathcal{A}_{i,t}\subseteq\mathcal{P}_{i,t}$ be the accepted subset. Define paper weights
\begin{equation}
\omega_p
=
\begin{cases}
1, & p \in \mathcal{A}_{i,t},\\
\kappa, & p \in \mathcal{P}_{i,t}\setminus\mathcal{A}_{i,t},
\end{cases}
\qquad \kappa\in(0,1).
\end{equation}
The restriction $\kappa<1$ means that accepted papers have a higher impact on identity evolution; $\kappa>0$ keeps the denominator of the identity update strictly positive even when none of a scientist's papers is accepted in a round.

Identity updates according to
\begin{equation}
\theta_{i,t+1}
=
(1-\lambda_i)\theta_{i,t}
+
\lambda_i
\left(
\frac{\sum_{p\in\mathcal{P}_{i,t}} \omega_p x_p}
{\sum_{p\in\mathcal{P}_{i,t}} \omega_p}
\right),
\end{equation}
with $\lambda_i\in(0,1]$ controlling identity plasticity. If $k_{i,t}=0$, $\theta_{i,t+1}=\theta_{i,t}$.

Review tastes then adjust toward the updated production identity,
\begin{equation}
\tau_{i,t+1}
=
(1-\eta_i)\tau_{i,t}
+
\eta_i \theta_{i,t+1},
\qquad \eta_i\in(0,1].
\end{equation}
Note the sequential updating: $\tau_{i,t+1}$ uses $\theta_{i,t+1}$, not $\theta_{i,t}$.

The overall procedure is summarized in Algorithm~\ref{algorithm:reciprocal_peer_review}.

\paragraph{\textbf{Emergent Concentration of Accepted Topics.}}
Although reviewer assignment is ex ante symmetric across scientists, heterogeneity in submission volumes translates through the reciprocity term $w_{j,t}=1+\rho k_{j,t}$ into a systematic bias in the composition of the reviewer pool: prolific scientists are disproportionately represented. Because the taste term $-\beta\|x_p-\tau_{j,t}\|^2$ penalizes papers distant from a reviewer's own taste, this over-representation translates directly into a bias in the scoring, and through rank-based acceptance, into a bias in which papers are accepted.

The model's central prediction is therefore the following: as $\rho$ increases, the distribution of accepted topics $\{x_p: \mathrm{Accept}_p=1, p\in\mathcal{P}_t\}$ concentrates around the tastes of high-submission scientists. This concentration arises immediately from the reviewer-assignment channel in round $t=1$ and is reinforced over time through the identity update. We emphasize that this effect does not require any institutional weighting of reviewer identities; it emerges endogenously from the interaction of reciprocal assignment and distance-based evaluation.

\paragraph{\textbf{Implementation.}}
We instantiate the model with $N=100$ scientists in a $d=2$ topic space and simulate $T=120$ submission rounds. We adopt a stylized two-type initialization that captures the distinction between an established research community and a broader periphery. A small clique of $N_P=5$ prolific scientists is initialized with tightly clustered identities, $\theta_{i,0} \sim \mathcal{N}(0, 0.2^2 I_2)$, while the remaining $95$ regular scientists are drawn from a wider prior, $\mathcal{N}(0, 1.5^2 I_2)$. Initial review tastes are set equal to initial identities for both groups, $\tau_{i,0}=\theta_{i,0}$. Submission rates are fixed over time and bimodal: prolific scientists draw $k_{i,t} \sim \mathrm{Poisson}(20)$ per round, while regular scientists draw $k_{i,t} \sim \mathrm{Poisson}(2)$. This asymmetry is the only exogenous source of reviewer-pool imbalance; the reviewer assignment is otherwise symmetric across scientists.

Each paper is drawn with within-author variability $\sigma_x = 0.1$ and receives $R=3$ reviewers, sampled sequentially without replacement from all non-author scientists with weights proportional to $\exp(-\delta \|x_p - \tau_{j,t}\|^2) \cdot w_{j,t}$, where $\delta = 0.5$. Review scores use baseline $\bar{s}=5$, quality penalty $\gamma = 0.2$, taste penalty $\beta = 1.5$, and noise $\sigma = 0.3$; the ratio $\gamma/\beta \approx 0.13$ places the model in a taste-inflected regime where the reviewer identity has a substantial influence relative to the intrinsic paper quality, consistent with critiques of subjective evaluation in peer review \citep{beygelzimer2023has}. Each round, the top $\pi = 0.25$ fraction of papers by mean score is accepted, with ties broken at random. Identity updates use rejected-paper weight $\kappa = 0.1$, with identity and taste plasticities $\lambda_i = \eta_i = 0.25$ uniform across scientists, although this design can be further complicated by introducing heterogeneity in plasticity, for example, senior scientists may exhibit lower research plasticity. To isolate the role of reciprocity, we vary $\rho \in \{0, 3, 15\}$: $\rho=0$ disables the reciprocal channel entirely (uniform reviewer availability, so the reviewer pool composition is determined only by topical matching), $\rho=3$ represents moderate reciprocity, and $\rho=15$ represents a strongly reciprocal system in which the reviewing load is tightly coupled to submission load. For each value of $\rho$ we run $25$ independent replicates with different random seeds and report the ensemble means with $\pm 1$ standard error bands. The primary outcome is the \emph{concentration of accepted topics around the establishment's tastes}: in each round we compute the mean Euclidean distance from each accepted paper's topic $x_p$ to the centroid of the prolific clique's tastes,
\begin{equation}
\bar{\tau}^{\mathrm{prol}}_t
=
\frac{1}{N_P} \sum_{j \in \mathrm{prolific}} \tau_{j,t}.
\end{equation}
Lower values of this distance indicate that the set of accepted papers is more tightly clustered around the tastes of prolific scientists, signaling that the review system is disproportionately rewarding topics aligned with the establishment.



\subsection{Subfield Concordance}
\label{sec:app_subfield_concordance}

We categorize computer science papers into three broad groups following standard field conventions from \texttt{csrankings.org}. The classification maps each of OpenAlex's hierarchical identifiers (\texttt{Txxxxx}: identifier name) to one of three labels: AI, System, Theory, using a manually curated concordance table (see below). Each paper is assigned the subfield label.

\vspace{0.5em}
\textbf{AI/Machine Learning -- AI}
\vspace{0.5em}

T10036: Advanced Neural Network Applications; T10100: Metaheuristic Optimization Algorithms Research; T10203: Recommender Systems and Techniques; T10211: Computational Drug Discovery Methods; T10249: Distributed Control Multi-Agent Systems; T10286: Information Retrieval and Search Behavior; T10320: Neural Networks and Applications; T10456: Multi-Agent Systems and Negotiation; T10462: Reinforcement Learning in Robotics; T10538: Data Mining Algorithms and Applications; T10586: Robotic Path Planning Algorithms; T10637: Advanced Clustering Algorithms Research; T10820: Fuzzy Logic and Control Systems; T10862: AI in cancer detection; T10906: AI-based Problem Solving and Planning; T11010: Logic, Reasoning, and Knowledge; T11273: Advanced Graph Neural Networks; T11303: Bayesian Modeling and Causal Inference; T11307: Domain Adaptation and Few-Shot Learning; T11347: Neural Networks Stability and Synchronization; T11512: Anomaly Detection Techniques and Applications; T11574: Artificial Intelligence in Games; T11612: Stochastic Gradient Optimization Techniques; T11644: Spam and Phishing Detection; T11652: Imbalanced Data Classification Techniques; T11689: Adversarial Robustness in Machine Learning; T11901: Bayesian Methods and Mixture Models; T11902: Intelligent Tutoring Systems and Adaptive Learning; T11975: Evolutionary Algorithms and Applications; T12016: Web Data Mining and Analysis; T12026: Explainable Artificial Intelligence (XAI); T12072: Machine Learning and Algorithms; T12128: AI in Service Interactions; T12380: Authorship Attribution and Profiling; T12535: Machine Learning and Data Classification; T12611: Neural Networks and Reservoir Computing; T12676: Machine Learning and ELM; T12761: Data Stream Mining Techniques; T12805: Cognitive Science and Mapping; T12814: Gaussian Processes and Bayesian Inference; T13062: Cognitive Computing and Networks; T13373: Data Mining and Machine Learning Applications; T13514: Experience-Based Knowledge Management; T13567: AI and Multimedia in Education; T13647: AI and Big Data Applications; T13693: Smart Systems and Machine Learning; T13702: Machine Learning in Healthcare; T13748: Advanced Statistical Modeling Techniques; T13832: Advanced Decision-Making Techniques; T13904: Artificial Intelligence Applications; T14174: Artificial Intelligence and Decision Support Systems; T14175: Intuitionistic Fuzzy Systems Applications; T14335: Educational Robotics and Engineering; T14347: Big Data and Digital Economy; T14351: Statistical and Computational Modeling; T14414: Artificial Intelligence in Education; T14435: Information Retrieval and Data Mining.

\vspace{0.5em}
\textbf{Computer Vision -- AI}
\vspace{0.5em}

T10052: Medical Image Segmentation Techniques; T10057: Face and Expression Recognition; T10331: Video Surveillance and Tracking Methods; T10444: Context-Aware Activity Recognition Systems; T10531: Advanced Vision and Imaging; T10601: Handwritten Text Recognition Techniques; T10627: Advanced Image and Video Retrieval Techniques; T10688: Image and Signal Denoising Methods; T10775: Generative Adversarial Networks and Image Synthesis; T10812: Human Pose and Action Recognition; T10824: Image Retrieval and Classification Techniques; T10828: Biometric Identification and Security; T11017: Chaos-based Image/Signal Encryption; T11019: Image Enhancement Techniques; T11105: Advanced Image Processing Techniques; T11165: Image and Video Quality Assessment; T11398: Hand Gesture Recognition Systems; T11439: Video Analysis and Summarization; T11448: Face recognition and analysis; T11605: Visual Attention and Saliency Detection; T11714: Multimodal Machine Learning Applications; T12357: Digital Media Forensic Detection; T12549: Image and Object Detection Techniques; T12874: Digital Imaging for Blood Diseases; T12923: Digital Image Processing Techniques; T13270: QR Code Applications and Technologies; T13579: Image and Video Stabilization; T14319: Currency Recognition and Detection; T14339: Image Processing and 3D Reconstruction.

\vspace{0.5em}
\textbf{Natural Language Processing -- AI}
\vspace{0.5em}

T10028: Topic Modeling; T10181: Natural Language Processing Techniques; T10215: Semantic Web and Ontologies; T10664: Sentiment Analysis and Opinion Mining; T11550: Text and Document Classification Technologies; T12262: Hate Speech and Cyberbullying Detection; T13083: Advanced Text Analysis Techniques; T13274: Expert finding and Q\&A systems; T13623: Linguistic Studies and Language Acquisition; T13629: Text Readability and Simplification.

\vspace{0.5em}
\textbf{Human-Computer Interaction+AI -- AI}
\vspace{0.5em}

T13851: Law, AI, and Intellectual Property.

\vspace{0.5em}
\textbf{Architecture and Systems -- System}
\vspace{0.5em}

T10054: Parallel Computing and Optimization Techniques; T10101: Cloud Computing and Resource Management; T10715: Distributed and Parallel Computing Systems; T10772: Distributed systems and fault tolerance; T10904: Embedded Systems Design Techniques; T10933: Real-Time Systems Scheduling; T11032: VLSI and Analog Circuit Testing; T12122: Physical Unclonable Functions (PUFs) and Hardware Security; T13608: Advanced Computational Techniques in Science and Engineering; T14067: Cloud Computing and Remote Desktop Technologies; T14098: Energy Efficiency in Computing.

\vspace{0.5em}
\textbf{Computer Networks -- System}
\vspace{0.5em}

T10080: Energy Efficient Wireless Sensor Networks; T10138: Network Traffic and Congestion Control; T10246: Mobile Ad Hoc Networks; T10273: IoT and Edge/Fog Computing; T10400: Network Security and Intrusion Detection; T10575: Wireless Communication Networks Research; T10579: Cognitive Radio Networks and Spectrum Sensing; T10711: Target Tracking and Data Fusion in Sensor Networks; T10714: Software-Defined Networks and 5G; T10742: Peer-to-Peer Network Technologies; T10796: Cooperative Communication and Network Coding; T10829: Interconnection Networks and Systems; T11158: Wireless Networks and Protocols; T11478: Caching and Content Delivery; T11498: Security in Wireless Sensor Networks; T11704: Mobile Crowdsensing and Crowdsourcing; T11896: Opportunistic and Delay-Tolerant Networks; T12131: Wireless Signal Modulation Classification; T12203: Mobile Agent-Based Network Management; T12216: Network Time Synchronization Technologies; T12326: Network Packet Processing and Optimization; T12564: Sensor Technology and Measurement Systems; T12801: Bluetooth and Wireless Communication Technologies; T12879: Distributed Sensor Networks and Detection Algorithms; T13038: Internet of Things and AI; T13541: IoT-based Control Systems; T13924: Internet of Things and Social Network Interactions; T14353: Wireless Sensor Networks for Data Analysis.

\vspace{0.5em}
\textbf{Computer Security -- System}
\vspace{0.5em}

T10237: Cryptography and Data Security; T10270: Blockchain Technology Applications and Security; T10734: Information and Cyber Security; T10764: Privacy-Preserving Technologies in Data; T10951: Cryptographic Implementations and Security; T11241: Advanced Malware Detection Techniques; T11424: Security and Verification in Computing; T11504: Advanced Authentication Protocols Security; T11598: Internet Traffic Analysis and Secure E-voting; T11614: Cloud Data Security Solutions; T11800: User Authentication and Security Systems; T12034: Digital and Cyber Forensics; T12479: Web Application Security Vulnerabilities; T13983: Cybersecurity and Information Systems; T14042: Technology and Security Systems.

\vspace{0.5em}
\textbf{Databases -- System}
\vspace{0.5em}

T10317: Advanced Database Systems and Queries; T11106: Data Management and Algorithms; T11181: Advanced Data Storage Technologies; T11813: Library Collection Development and Digital Resources; T11937: Research Data Management Practices; T12601: Web Applications and Data Management; T12863: Web and Library Services; T13166: Information Science and Libraries; T13673: Library Science and Information; T14330: Library Science and Information Systems.

\vspace{0.5em}
\textbf{Software Engineering - System}
\vspace{0.5em}

T10260: Software Engineering Research; T10430: Software Engineering Techniques and Practices; T10639: Advanced Software Engineering Methodologies; T10679: Service-Oriented Architecture and Web Services; T10743: Software Testing and Debugging Techniques; T11450: Model-Driven Software Engineering Techniques; T11675: Open Source Software Innovations; T12127: Software System Performance and Reliability; T12423: Software Reliability and Analysis Research; T12799: Mobile and Web Applications; T13197: Spreadsheets and End-User Computing.

\vspace{0.5em}
\textbf{Theory -- Theory}
\vspace{0.5em}

T10020: Quantum Information and Cryptography; T10126: Logic, programming, and type systems; T10142: Formal Methods in Verification; T10374: Advanced Graph Theory Research; T10388: Advanced Steganography and Watermarking Techniques; T10545: Optimization and Variational Analysis; T10682: Quantum Computing Algorithms and Architecture; T10720: Complexity and Algorithms in Graphs; T10792: Matrix Theory and Algorithms; T10848: Advanced Multi-Objective Optimization Algorithms; T10901: Advanced Data Compression Techniques; T11063: Rough Sets and Fuzzy Logic; T11125: Petri Nets in System Modeling; T11130: Coding theory and cryptography; T11187: Nonlinear Dynamics and Pattern Formation; T11269: Algorithms and Data Compression; T11321: Error Correcting Code Techniques; T11435: Polynomial and algebraic computation; T11567: semigroups and automata theory; T11596: Constraint Satisfaction and Optimization; T11693: Cryptography and Residue Arithmetic; T11697: Numerical Methods and Algorithms; T11727: Advanced Algebra and Logic; T12002: Computability, Logic, AI Algorithms; T12100: Advanced Mathematical Modeling in Engineering; T12162: Cellular Automata and Applications; T12288: Optimization and Search Problems; T12292: Graph Theory and Algorithms; T12536: Topological and Geometric Data Analysis; T12541: Graph Labeling and Dimension Problems; T12794: Adaptive Dynamic Programming Control; T12809: Contact Mechanics and Variational Inequalities; T13182: Quantum-Dot Cellular Automata; T13286: Modeling and Simulation Systems; T13523: Mathematics, Computing, and Information Processing.

\subsection{Replication of Cyclical Exhaustion in Structural Biology}
\label{sec:bio}

A natural question is whether our main findings about cyclical exhaustion reflect general properties of knowledge production or features peculiar to computer science. Computer science is unusually shaped by industry: well-resourced firms employ and fund a substantial share of active researchers, so the agenda driving publication may be as much corporate as epistemic. Consolidation in such a field might therefore index competitive economic pressure rather than any fundamental dynamic of collective inquiry. To assess this, we replicate our main results in a second field, structural biology. We collected all structural biology papers indexed in OpenAlex from 2010 to 2025 and applied our measure of new/active ideas in the main text without modification. We recover the same pattern of cyclical exhaustion; in what follows we consider the field-specific mechanisms that may account for it.

Structural biology is widely regarded as having undergone two paradigm shifts in recent decades. The first is the cryo-electron microscopy (cryo-EM) "resolution revolution": although unlike computer science, cryo-EM had been under development since the 1970s \textit{without} a single defining founding paper, advances in direct electron detectors and image processing brought it to near-atomic resolution in the 2010s and established it as a mainstream method of the field \citep{kuhlbrandt2014resolution}. The second is AI-based structure prediction, inaugurated by AlphaFold, which made high-confidence atomic models available for essentially any protein sequence \citep{subramaniam2022paradigm}. That these two episodes constitute the field's defining transitions is not our interpretation but the field's own \citep{subramaniam2026cryo, kuhlbrandt2014resolution, subramaniam2022paradigm}. Both episodes were recognized by Nobel Prizes in Chemistry, in 2017 and 2024 respectively.

As shown in Supplementary Fig.~\ref{fig:crossfield_robustness}, our measure of new ideas recovers this structure precisely: we observe two inverted-U shapes using exactly the measurement described in the main text. The first inverted U is the more pronounced of the two and aligns with the maturation of cryo-EM. Notably, new ideas begin to decline before the field's moment of bright recognition (the \textit{Nature Methods} designation as ``Method of the Year'' in late 2015) and continue to decline thereafter, suggesting that formal recognition marks the consolidation of a paradigm rather than its generative frontier. A second rise begins in late 2018, coinciding with AlphaFold's first appearance at CASP13, but the resulting peak is markedly weaker than the first. Domain experts attribute this asymmetry to breadth of adoption: cryo-EM reshaped experimental practice across the entire structural biology community, whereas AlphaFold's early impact was concentrated in the computational subcommunity. These patterns mirror the dynamics of the "carrying capacity" of active ideas: by the time cryo-EM received major recognition, the growth of active ideas had already slowed and even begun to decline later, whereas following the release of AlphaFold, the number of active ideas began to grow again. This difference also speaks to the concern that partly motivated the replication. Cryo-EM emerged from academic instrumentation development rather than from a single firm's research agenda, yet it produces the same cyclical signature as AlphaFold and AI, the paradigm shifts (partly) driven by the industry. The pattern therefore does not appear to depend on corporate sponsorship of the underlying advance. 

Structural biology differs from AI in one important respect. Its curve of the second rise of active ideas, driven by AlphaFold, is much stronger than the corresponding growth in new ideas. The likely explanation is that AlphaFold generated new concepts only within a relatively small computational subcommunity, while its broader effect was to revive ideas that had nearly gone dormant: by cracking long-standing protein structures that could not even be solved by cryo-EM, it reopened research around them without introducing new concepts of its own. We also note a third upturn in new ideas beginning in 2024, for which no established explanation is available; we speculate that it reflects the diffusion of generative AI and LLMs, which appear to have injected new vitality not only into structural biology but also across the sciences.

\newpage

\begin{algorithm}[p]
\caption{\textbf{Reciprocal Peer Review and Identity Dynamics}}
\label{algorithm:reciprocal_peer_review}

Fix $N$, $T$, $d$, $R$, $\pi=0.25$, and $\delta,\beta,\gamma,\bar{s},\sigma,\rho,\kappa,\sigma_x,\{\mu_i\},\{\lambda_i,\eta_i\}$\;
Initialize $\theta_{i,0}$ from the type-specific prior and set $\tau_{i,0}=\theta_{i,0}$ for $i=1,\dots,N$\;

\For{each submission round $t$}
{
    Initialize $\mathcal{P}_t\leftarrow\emptyset$\;
    \For{each scientist $i = 1,\dots,N$}
    {
        Draw $k_{i,t}\sim\mathrm{Poisson}(\mu_i)$\;
        \For{$m=1$ to $k_{i,t}$}
        {
            Draw $x_p\sim\mathcal{N}(\theta_{i,t},\sigma_x^2 I_d)$\;
            Add $p=(i,x_p)$ to $\mathcal{P}_t$\;
        }
    }
    Compute $w_{j,t}=1+\rho k_{j,t}$ for all $j$\;

    \For{each paper $p \in \mathcal{P}_t$}
    {
        Sample $\mathcal{R}(p)$ of size $R$ from
        $\{j\neq i(p)\}$ with weights $\psi_{j,p,t}\propto\exp(-\delta\|x_p-\tau_{j,t}\|^2)\,w_{j,t}$\;
        \For{$j \in \mathcal{R}(p)$}
        {
            Draw $\varepsilon_{jp}\sim\mathcal{N}(0,\sigma^2)$; compute $s_{jp}$\;
        }
        Compute $S_p$\;
    }
    Set $\bar{S}_t=$ the $(1-\pi)$-quantile of $\{S_p\}$; set $\mathrm{Accept}_p=\mathbf{1}\{S_p\ge\bar{S}_t\}$ (ties broken at random)\;

    \For{each scientist $i = 1,\dots,N$}
    {
        \eIf{$k_{i,t}>0$}
        {
            Set $\omega_p=1$ if $p\in\mathcal{A}_{i,t}$, else $\omega_p=\kappa$\;
            Update $\theta_{i,t+1}$\;
        }{
            $\theta_{i,t+1}\leftarrow\theta_{i,t}$\;
        }
        Update $\tau_{i,t+1}$\;
    }
}
\end{algorithm}

\clearpage
\newpage

\section{Robustness Checks}
\label{sec:robustness}


\subsection{Gini Robustness Across \texorpdfstring{$k$}{k}}
\label{sec:app_attention_robustness}

To verify the robustness of attention inequality results, we repeat the measurement under multiple values of $k$ (see Supplementary Section~\ref{sec:app_attention}), the number of keywords extracted per paper.

\begin{figure}[H]
    \centering
    \includegraphics[width=0.6\textwidth]{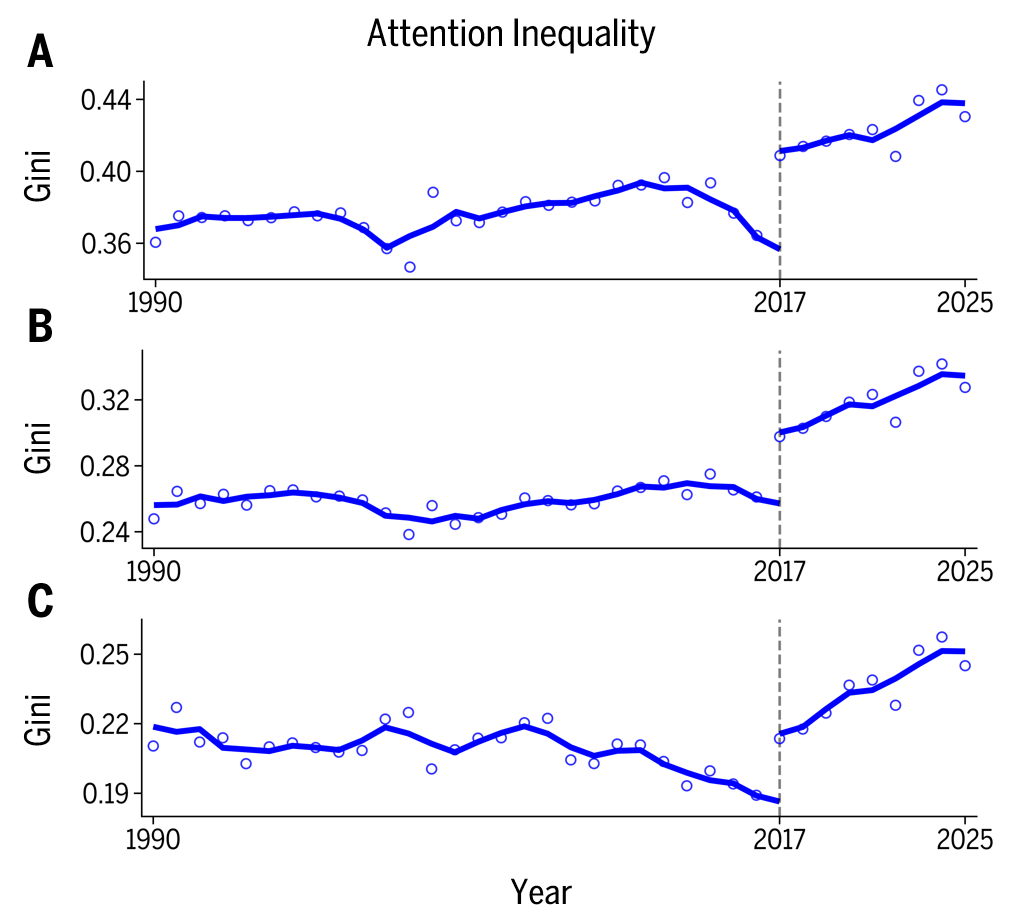}
\caption{\textbf{Robust results on attention concentration.}
Gini coefficient under different TF-IDF keyword extraction parameters. Panels A, B, and C correspond to $k = 5$, $k = 10$, and $k = 15$, respectively. Lines are fitted using a three-year rolling average.}
    \label{fig:attention_robustness}
\end{figure}

Supplementary Fig.~\ref{fig:attention_robustness} panels A, B, and C present the results for $k = 5$, $k = 10$, and $k = 15$, where panel B corresponds to the result reported in the paper. Due to differences in parameter settings, the range of the Gini coefficient varies across specifications. However, the conclusion that the Gini coefficient rises sharply after 2017 is robust across all these settings.


\subsection{Idea-Extraction Robustness Across NLTK / spaCy / Noun-Phrases}
\label{sec:app_idea_robustness}

\begin{figure}[H]
    \centering
    \includegraphics[width=1\textwidth]{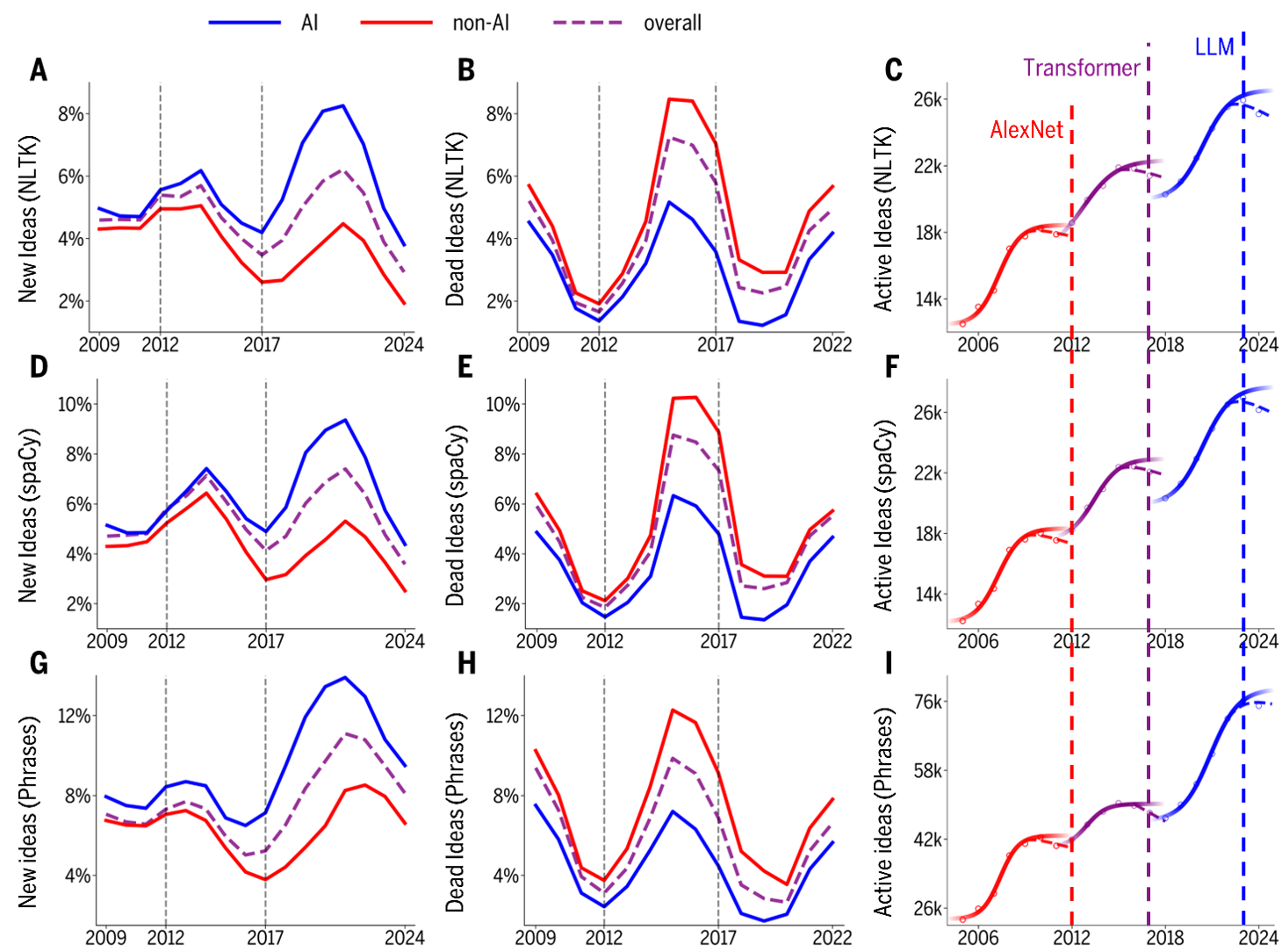}
\caption{\textbf{Robust results on idea representation.}
Robustness of idea lifecycle metrics across idea representation methods. Conclusions are robust.}
    \label{fig:idea_robustness}
\end{figure}

The idea life-cycle metrics reported in the paper (Fig. 2) are computed using NLTK-extracted nouns from the dataset. To verify the robustness of these findings to the choice of idea representation methods, we replicated all analyses using spaCy as an alternative extraction pipeline, as well as using ``noun phrases'' extracted by spaCy, that is, contiguous spans headed by a noun and including its left-side modifiers (determiners, adjectives, etc.), for instance, ``the quick brown fox''. All results are robust. The cyclical pattern of new ideas and dead ideas, and the S-shaped growth of the active ideas are robustly observed across any idea representation methods (Supplementary Fig.~\ref{fig:idea_robustness}).

\subsection{Author-Migration Regression Tables Across Alternative Status Measures}
\label{sec:app_migration_regression}

As shown in Supplementary Tables~\ref{tab:migration_ai} and~\ref{tab:migration_ai2}, we use two different representations of author status (publication-based and citation-based). Across the regression specifications, we find that the following conclusions are robust:

\begin{itemize}
    \item Mid-career scientists show the strongest propensity to transition into AI: the squared terms of both age and prior status are negative and significant, indicating an inverted U-shaped relationship.
    \item In years when authors publish a great deal, they are less likely to switch (the coefficient on current-year publications is negative and significant).
    \item The longer authors have stayed in their current field, the less likely they are to switch (the coefficient on tenure in the current field is negative and significant).
    \item Authors with longer publication gaps are more likely to switch (the coefficient on unpublished years is positive and significant).
\end{itemize}


\clearpage
\section{Supplementary Figures and Tables} \label{sec:si_figures_tables}

\phantomsection
\addcontentsline{toc}{subsection}{Supplementary Fig.~1: Robust Results on Attention Concentration}

\clearpage
\phantomsection
\addcontentsline{toc}{subsection}{Supplementary Fig.~2: Robust Results on Idea Representation}
\clearpage

\phantomsection
\addcontentsline{toc}{subsection}{Supplementary Fig.~3: Illustration of the Four-stage Spiral of Technological Revolution}
\begin{figure}[H]
    \centering
    \includegraphics[width=0.6\textwidth]{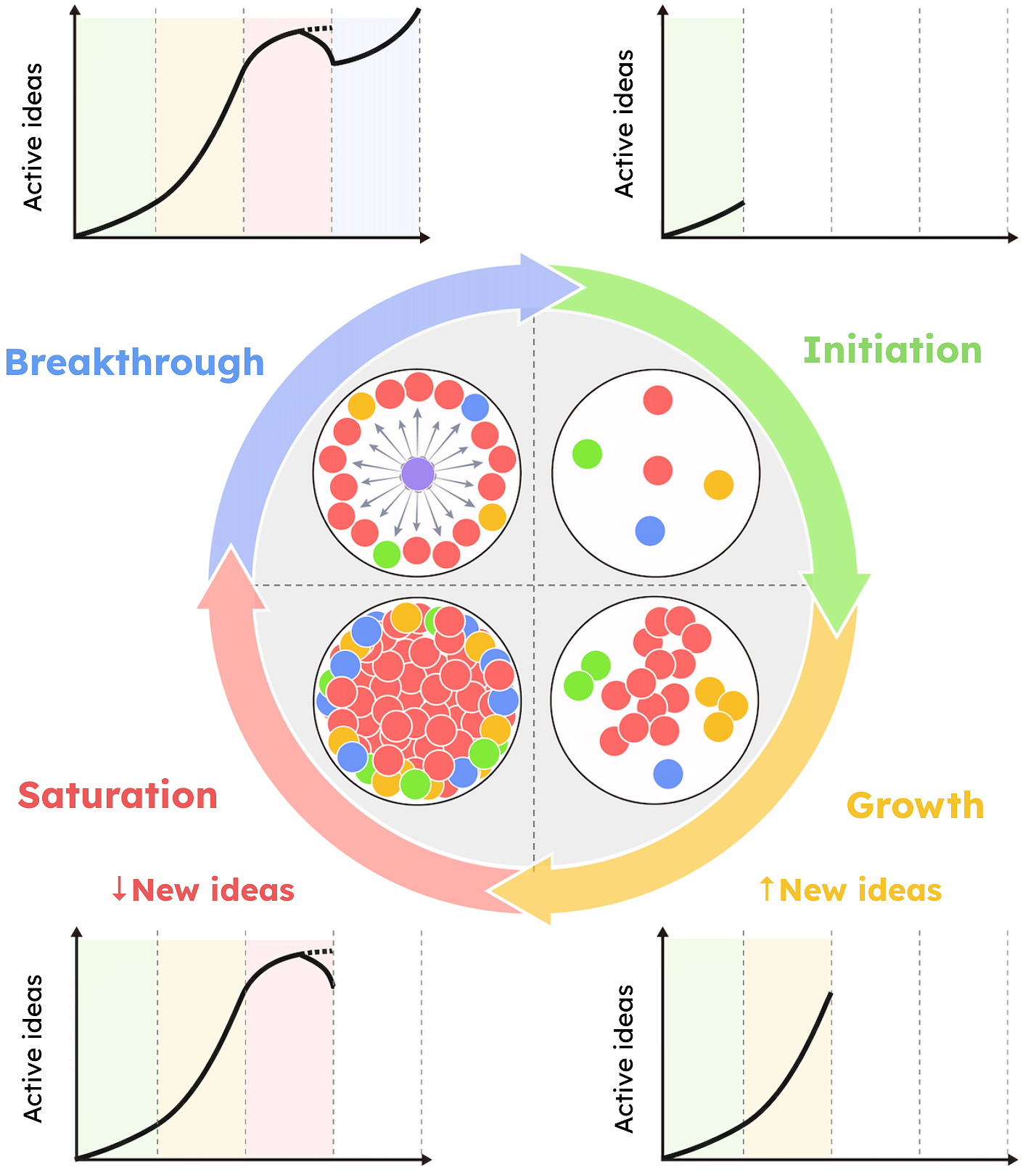}
\caption{\textbf{Illustration of the four-stage spiral of technological revolution.}
Across 36 years of AI development including three major technological advances (AlexNet, the Transformer, and LLMs), we repeatedly observe the same pattern, a \textbf{four-stage spiral of technological revolution}: \emph{initiation}, where a new paradigm has just emerged; \emph{growth}, where the S-curve reaches its steepest slope and new ideas proliferate; \emph{saturation}, where the idea population calcifies and may even deteriorate; and finally \emph{breakthrough}, a largely unpredictable transformation that lifts the carrying capacity and restarts the cycle.}
    \label{fig:illu}
\end{figure}
\clearpage

\phantomsection
\addcontentsline{toc}{subsection}{Supplementary Fig.~4: Cyclical Exhaustion in Structural Biology}
\begin{figure}[H]
    \centering
    \includegraphics[width=1\textwidth]{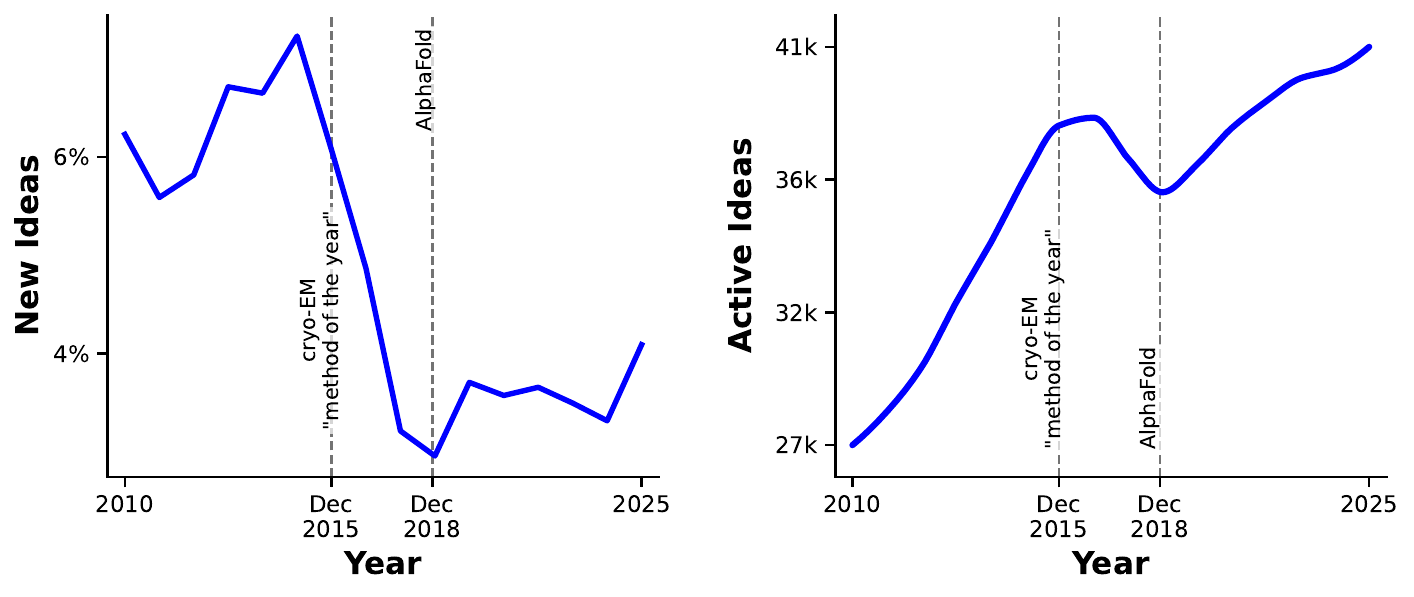}
\caption{\textbf{Cyclical exhaustion in structural biology.}
Left: Recognition of cryo-EM came later than the peak in new ideas. AlphaFold produced a second, smaller peak. Right: Active ideas follow the same growth pattern as in computer science: each breakthrough raises the carrying capacity of idea space, which is then repeatedly exhausted.
}
    \label{fig:crossfield_robustness}
\end{figure}
\clearpage

\phantomsection
\addcontentsline{toc}{subsection}{Supplementary Table~1: Determinants of Author Migration to AI (Publications-based Status)}
\begin{table}[H]\centering
\caption{Determinants of author migration to AI (using yearly publication percentile to represent prior status)}
\label{tab:migration_ai}
\begin{tabular}{lcc}
\toprule
& (1) & (2) \\
& Logit & Logit \\
\midrule
Age 
    & 0.025*** 
    & 0.073*** \\
    & (0.000) 
    & (0.001) \\
Age$^2$ 
    & 
    & -0.001*** \\
    & 
    & (0.000) \\
Prior Status
    & 0.265*** 
    & 1.083*** \\
    & (0.007) 
    & (0.023) \\
Prior Status$^2$ 
    & 
    & -0.658*** \\
    & 
    & (0.022) \\
Curr. Pub. (pct) 
    & -0.397*** 
    & -0.433*** \\
    & (0.007) 
    & (0.007) \\
Tenure in Curr. Field 
    & -0.172*** 
    & -0.193*** \\
    & (0.001) 
    & (0.001) \\
Unpublished Years 
    & 0.056*** 
    & 0.057*** \\
    & (0.001) 
    & (0.001) \\
\midrule
Year FE & Yes & Yes \\
Observations & 3,721,331 & 3,721,331 \\
Pseudo $R^2$       & 0.041 & 0.044 \\
\bottomrule
\multicolumn{2}{p{0.5\linewidth}}{\footnotesize
\textit{Notes:} Standard errors clustered at the author level. \newline
*** $p<0.001$, ** $p<0.01$, * $p<0.05$.
}
\end{tabular}
\end{table}
\clearpage
\phantomsection
\addcontentsline{toc}{subsection}{Supplementary Table~2: Determinants of Author Migration to AI (Citations-based Status)}
\begin{table}[H]\centering
\caption{Determinants of author migration to AI (using yearly citation percentile to represent prior status)}
\label{tab:migration_ai2}
\begin{tabular}{lcc}
\toprule
& (1) & (2) \\
& Logit & Logit \\
\midrule
Age 
    & 0.021*** 
    & 0.056*** \\
    & (0.000) 
    & (0.001) \\
Age$^2$ 
    & 
    & -0.001*** \\
    & 
    & (0.000) \\
Prior Status 
    & 0.395*** 
    & 0.490*** \\
    & (0.008) 
    & (0.029) \\
Prior Status$^2$ 
    & 
    & -0.150*** \\
    & 
    & (0.027) \\
Curr. Pub. (pct) 
    & -0.398*** 
    & -0.391*** \\
    & (0.006) 
    & (0.007) \\
Tenure in Curr. Field 
    & -0.183*** 
    & -0.187*** \\
    & (0.001) 
    & (0.001) \\
Unpublished Years 
    & 0.056*** 
    & 0.051*** \\
    & (0.000) 
    & (0.000) \\
\midrule
Year FE & Yes & Yes \\
Observations & 3,721,331 & 3,721,331 \\
Pseudo $R^2$       & 0.042 & 0.043 \\
\bottomrule
\multicolumn{2}{p{0.5\linewidth}}{\footnotesize
\textit{Notes:} Standard errors clustered at the author level. \newline
*** $p<0.001$, ** $p<0.01$, * $p<0.05$.
}
\end{tabular}
\end{table}

\end{document}